\documentclass[runningheads]{llncs}
\usepackage{eccv}
\usepackage{eccvabbrv}
\usepackage{graphicx}
\usepackage{booktabs}
\usepackage[accsupp]{axessibility}  
\usepackage{times}
\usepackage{epsfig}
\usepackage{amsmath}
\usepackage{amssymb}
\usepackage{color}
\usepackage{comment}
\usepackage{lipsum}
\usepackage{enumitem}
\usepackage{float}
\usepackage{makecell}
\usepackage{pifont}
\usepackage{animate}
\usepackage[normalem]{ulem}
\usepackage{cite}
\usepackage{tikz}
\usetikzlibrary{positioning}
\usepackage{pgfplots}
\usepackage{multirow}

\usepackage{hyperref}
\usepackage{orcidlink}

\definecolor{darkgreen}{RGB}{30,150,30}
\definecolor{darkred}{RGB}{150,30,30}
\definecolor{darkblue}{RGB}{30,30,150}
\definecolor{blue}{RGB}{0,0,255}
\definecolor{magenta}{RGB}{175,0,175}
\definecolor{teal}{RGB}{0,128,128}
\definecolor{orange}{RGB}{255,165,0}
\definecolor{ROC1}{RGB}{233,116,81}
\definecolor{ROC2}{RGB}{30,175,30}
\definecolor{ROC3}{RGB}{0,128,128}

\usepackage[capitalize]{cleveref}
\crefname{section}{Sec.}{Secs.}
\Crefname{section}{Section}{Sections}
\Crefname{table}{Table}{Tables}
\crefname{table}{Tab.}{Tabs.}

\begin{document}
\def\frameFeat{\mathbf{d}_f} 
\newcommand\clip[2]{\mathcal{C}_{\text{ID}_{#1}\rightarrow \text{ID}_{#2}}}
\newcommand\shuffledclip[2]{\overset{\sim}{\mathcal{C}}_{\text{ID}_{#1}\rightarrow \text{ID}_{#2}}}

\title{Avatar Fingerprinting for Authorized Use of \\Synthetic Talking-Head Videos}
\author{Ekta Prashnani \and
Koki Nagano \and
Shalini De Mello \and
David Luebke \and
Orazio Gallo
}

\titlerunning{Avatar Fingerprinting for Authorized Use of Synthetic Talking-Head Videos}
\authorrunning{Prashnani, Nagano, De Mello, Luebke, and Gallo.}
\institute{NVIDIA\\\email{{\{eprashnani, knagano, shalinig, dluebke, ogallo\}}@nvidia.com}\\}

\maketitle

\begin{abstract}
\vspace{-0.1in}
Modern avatar generators allow anyone to synthesize photorealistic real-time talking avatars, ushering in a new era of avatar-based human communication, such as with immersive AR/VR interactions or videoconferencing with limited bandwidths.
Their safe adoption, however, requires a mechanism to verify if the rendered avatar is trustworthy:
does it use the appearance of an individual without their consent?
We term this task \emph{avatar fingerprinting}.
To tackle it, we first introduce a large-scale dataset of real and synthetic videos of people interacting on a video call, where the synthetic videos are generated using the facial appearance of one person and the expressions of another.
We verify the identity driving the expressions in a synthetic video, by learning motion signatures that are independent of the facial appearance shown.
Our solution, the first in this space, achieves an average AUC of $0.85$.
Critical to its practical use, it also generalizes to new generators never seen in training (average AUC of $0.83$).
The proposed dataset and other resources can be found at: \url{https://research.nvidia.com/labs/nxp/avatar-fingerprinting/.}
\keywords{\hspace{-0.02in}Synthetic Media Verification \and Avatar Generators \and Avatar Attribution}
\end{abstract}

\section{Introduction}\label{sec:intro}

Recent digital avatar generators have fueled a myriad of computer vision and graphics applications, allowing anyone to synthesize real-time photorealistic personas.
Major companies are now supporting avatar-driven remote interactions over immersive AR/VR (\eg Meta's Pixel Codec Avatar~\cite{ma2021pixel}, Apple's Vision Pro Persona~\cite{visionpro}) or video conferencing (\eg NVIDIA's MAXINE~\cite{maxine}, Microsoft's Teams~\cite{msteams}), and selfie filters for altering and enhancing appearance (\eg by Snap and Tiktok).
While the avatar generation technology today is still young, the legitimate use of synthetic avatars will be ubiquitous in the future.
Without proper guardrails, this poses a real risk of unauthorized use and large-scale spread of visual disinformation.
To ensure the safe use of such a technology, the relevant question is no longer whether the content is ``real'' or not---since, by design, the videos and avatars are all \textit{synthetic}---but rather, whether the synthetically-generated videos and avatars are ``trustworthy" or not.

\begin{figure}
    \centering\includegraphics[width=0.72\columnwidth]{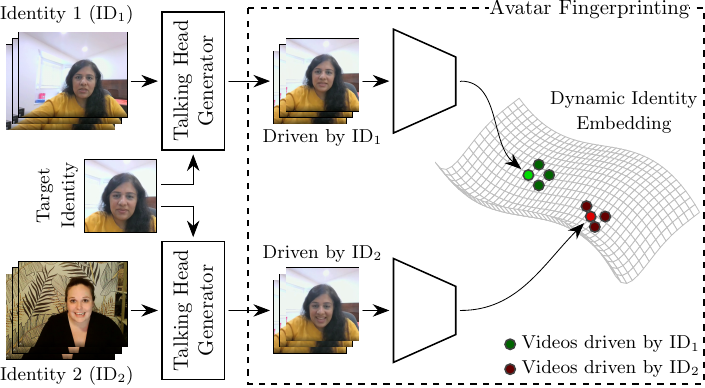}
    \vspace{-0.1in}
    \caption{Talking-head avatar generators can synthesize realistic videos of a target identity from driving videos of different identities.
    Our method extracts appearance-agnostic temporal facial features and learns an embedding in which the synthetic videos driven by one identity fall close to each other and far from those driven by other identities, regardless of the appearance of the synthetic video.
    By comparing distances in the embedding space, we evaluate whether an avatar is driven by an authorized identity or not.
    During evaluation, we only rely on the synthetic videos as input, without requiring any prior knowledge about the driving identity.}
    \label{fig:teaser}
    \vspace{-0.15in}
\end{figure}

When video conferencing, for instance, a synthetic video portrait generator can be used to save valuable bandwidth by reconstructing a synthetic avatar of the sender at the receiver's end, using only a frame of the target identity and a compact representation of the speaker's facial motion.
To ensure the authorized use of such synthetically-generated videos, we want to verify if the driving identity behind a synthetic video (ID$_1$ or ID$_2$ in Figure~\ref{fig:teaser}) is authorized to control the likeness, or the appearance, of the synthetic video (target identity in Figure~\ref{fig:teaser}). 
Crucially, we want to only leverage the synthetic video avatars to do so.
We call this novel task \emph{avatar fingerprinting}.

We leverage a simple but fundamental observation: facial motions tend to be idiosyncratic, that is, individuals talk and emote in unique ways.
For instance, someone may raise one of her eyebrows more than the other, or smile more while talking.
These ``dynamic identity signatures''~\cite{OToole_2002_CogSci} have been shown to carry enough information for humans to recognize other individuals, \emph{even when the physical appearance of their face is altered}~\cite{OToole_2002_CogSci,Hill_2001_CurrentBio,Knappmeyer_2001_JoV}.
This makes them attractive for our task, as they can be derived solely from the driving identity of a talking-head video, regardless of the appearance.

Fortunately, modern avatar generators are becoming increasingly accurate at capturing the facial motion of a person and rendering it onto a target identity.
As a first solution to avatar fingerprinting, we then propose to estimate a dynamic identity signature for an identity from the synthesized videos they drove---regardless of the target identities shown.
Specifically, we extract facial landmarks and their temporal dynamics from the video.
We then introduce a novel contrastive loss to learn a \emph{dynamic identity embedding}, a space where the dynamic identity signatures of a driving identity across multiple videos and target identities are close to each other, and far from those of other driving identities (see Figure~\ref{fig:teaser}).
We show that this approach, albeit straightforward, is robust and generalizes to generators not seen in training.

Avatar fingerprinting is a new task, and no existing datasets serve its training and validation requirements.
There are two key requirements from a dataset.
First, we need videos of multiple subjects delivering both scripted and free-form monologues, captured under realistic conditions, such as varying video quality and gaze direction.
This allows us to assess if a model leverages talking styles, and not the specific choice of words.
Second, to evaluate if the model learns to extract dynamic identity signatures effectively from synthetic videos, we need synthetic talking-head videos for the case in which the driving and target identities are different (cross-reenactment), and that in which they match (self-reenactment).
Unfortunately, existing datasets of real videos only fulfill a subset of these requirements (Table~\ref{table:dataset_features} and Section~\ref{sec:dataset}).
Further, no existing dataset of synthetic videos contain \textit{both} self- and cross-reenactments per subject or use the state-of-the-art talking-head generation technology.
To foster research in this new domain, then, we introduce the \uline{NV}IDIA \uline{Fa}c\uline{i}al \uline{R}eenactment (NVFAIR) dataset, containing real and synthetic face reenactment talking-head videos (Figure~\ref{fig:dataset}).
Our dataset, which includes ethnically diverse subjects, provides $10\times$ more synthetic facial reenactments than the second largest dataset, for a total of over $650,000$ synthetic videos.
It is also the only one using multiple state-of-the-art face reenactment generators~\cite{wang2021facevid2vid,wang2022latent,tps2022}, or to provide cross-reenactments driven by all identities, which is critical for training and evaluating avatar fingerprinting algorithms.

\vspace{0.1in}
\noindent In summary:
\begin{itemize}
    \item We introduce the novel task of avatar fingerprinting, which focuses on verifying the driving identity of synthetic talking-head avatars, rather than classifying them as real or synthetic (by design, all inputs to our model are synthetic).
    \item We release the first large-scale dataset of subjects delivering scripted and natural monologues, complete with self- and cross-reenactment videos synthesized with multiple state-of-the-art generators.
    \item We propose a solution for this novel task in the context of video conferencing by extracting person-specific motion signatures, and demonstrate its robustness to various distortions and avatar generators not seen in training.
\end{itemize}

\section{Related Work}\label{sec:related}
Our proposed avatar fingerprinting task aims to verify authorized use of synthetic avatars: this is fundamentally a different problem than traditional forensics research where one aims to detect synthetic media (\eg deepfake detection) or actively mark synthetic content.
In our case, by design, the content being evaluated is \textit{always synthetic}: we aim to evaluate its authorized use.
Since this is a novel task, no methods exist to directly address it. 
No methods currently exist to directly address this novel task.
Here we discuss the related areas of research.

\vspace*{-0.1in}
\paragraph{Learning-based Attribution of Synthetic Media.}
Learning-based approaches have been used to identify the origin of synthetic media, or to determine if it has been manipulated or altered in some way.
Previous work used a pre-trained GAN generator to attribute a synthesized image to its generator via GAN inversion by leveraging the fact that a real image is less invertible~\cite{Albright_2019_CVPR_Workshops,Karras2019stylegan2,ganscanner2021}. 
Yet other works focus on attributing other forms of synthetic media, such as text~\cite{munir2021through}.
In contrast, our focus is to attribute a talking-head avatar to the identity driving it, regardless of the appearance of the avatars.
Some existing works learn fingerprints associated with cameras to determine whether a video is manipulated~\cite{Verdoliva_2019_CVPR_Workshops}, or embed watermarks into images and videos~\cite{fridrich_2009,NIPS2017_838e8afb,2019stegastamp,Luo_2020_CVPR}, which are also shown to transfer to GAN-generated images~\cite{yu2021artificial}. 
Subsequent research introduced a watermark-based conditional GANs for scalable fingerprinting~\cite{yu2022responsible}. 
Our method, in contrast, is a \textit{passive} technique that does not rely on active watermarks.

\vspace*{-0.1in}
\paragraph{Deepfake Detection Based on Identity-Specific Features.}
Deepfake detection (``is a video synthetic?'') and avatar fingerprinting (``whose identity is used to generate this synthetic video?'') are fundamentally different tasks.
Most existing solutions for deepfake detection train a real-vs-synthetic classifier~\cite{Zhe21, sun2022dual, Hal21, li2022wavelet, ge2022deepfake, passos2022review}, and therefore cannot be adapted to avatar fingerprinting (where all inputs are synthetic). 
However, a specific class of detectors leverage identity-specific features to detect synthetic videos by posing the detection problem as an identity-recognition problem.
In our experiments, we evaluate some of these methods as baselines.
Specifically, Agarwal~\etal exploit person-specific patterns in facial expressions to detect fake videos~\cite{Agarwal_2019_CVPR_Workshops}.
ID-Reveal used facial shapes and motions encoded in a low-dimensional space of a 3D morphable model~\cite{3dmm1999} to handle both face-swapping and face-reenactment deepfakes~\cite{Cozzolino_2021_ICCV}. 
Other works explored soft-biometric approaches such as leveraging vocal mannerisms~\cite{Bohacek2022}, phoneme-viseme consistencies~\cite{Agarwal2020PhonemeViseme}, word-facial expression consistencies~\cite{Agarwal_2023_WACV, Agarwal2020DetectingDV}, and dynamics of ears~\cite{Agarwal_2021_CVPR}. 
While many of these works need person-specific training, previous works~\cite{TI2Net2023,Agarwal2020DetectingDV,Cozzolino_2021_ICCV,Cozzolino2022POI} extended this idea to train a CNN-based detector using a large-scale in-the-wild video data~\cite{Chung18b} and variants of contrastive learning~\cite{facenet2015,wang2019multi,contrastiveloss2006,Prannay2020contrastive}.
Agarwal~\etal combined static facial appearance using a facial recognition model and dynamic facial behaviors using a CNN, and showed that this approach is effective for detecting face-swap deepfakes~\cite{Agarwal2020DetectingDV}. 
Yet another line of research has explored temporal inconsistencies of face identities within a video~\cite{TI2Net2023}, and identity inconsistencies of inner and outer face regions~\cite{dong2022ict}.
Our experiments show that for avatar fingerprinting, such features that are designed to distinguish real from synthetic videos are not reliable, since we only have access to synthetic videos.   

\vspace*{-0.1in}
\paragraph{Dynamic Facial Identity Signatures.}
Cognitive scientists have studied the impact of ``dynamic facial identity signatures'' (\ie, characteristic or identity-specific movements of a face) for identity recognition for humans~\cite{OToole_2002_CogSci}.
In one experiment, scientists projected facial animations generated by human actors onto an average head and found that subjects discriminated between individuals based solely on facial motion~\cite{Hill_2001_CurrentBio}.
In another, subjects correctly attributed animations of synthetic faces to their morphed versions~\cite{Knappmeyer_2001_JoV}.
While these studies point to the existence of ``dynamic facial identity signatures'' that humans rely on, ours is the first method that isolates these from videos.

\vspace{-0.1in}
\paragraph{Talking-Head Datasets and Generators.}
Existing talking-head datasets include those that contain only real videos showing a variety of emotions ~\cite{livingstone2018ryerson,cao2014crema,wang2020mead}, and others that also contain synthetic videos~\cite{he2021forgerynet,roessler2019faceforensicspp,Celeb_DF_cvpr20,fox2021,kwon2021kodf,Dol20,Jia20}.
These datasets cater to traditional forensics and facial analysis tasks.
Therefore, they do not contain self- \textit{and} cross-reenactments driven by multiple identities, as well as scripted and free-form monologues across diverse capture settings.
There novel requirements posed by the avatar fingerprinting task motivate us to design our own dataset.
We focus specifically on face-reenactment talking-head avatar generators for synthetic video generation---this class of generators are the most relevant to AR/VR interactions, video conferencing, and several other applications~\cite{ma2021pixel,maxine,msteams,myheritage,heygen}---and combine various modes of human expression for capturing real videos.
Given a target facial image and a driving video, these generators reenact the target image using the facial expressions and head pose from the driving video~\cite{wang2022latent,hong2022depth,Siarohin_2019_NeurIPS,Zakharov_2019_ICCV,Zakharov20,Khakhulin2022ROME,Drobyshev22MP,wang2021facevid2vid,tps2022,mallya2022implicit}. 
Another class of talking-head generators use person-specific models~\cite{kim2018deep} and some models aim to preserve the style of the target identity in the synthesized video~\cite{kim2019neural}.
However, these models require person-specific training, making them difficult to scale.

\section{Terminology}\label{sec:terminology}

We seek to verify the trustworthiness of a synthesized talking-head video, termed \emph{target video}.
We assume that an avatar-generation tool (\eg,~\cite{wang2021facevid2vid}) created it by animating an image (\emph{target image}) using the expressions and head poses obtained from another video, the \emph{driving video}.
We call \emph{driving identity} the identity of the person in the driving video, and \emph{target identity} the identity of the person in the target image.
When driving and target identities match, the target video is a \emph{self-reenactment}, while the case of a driving identity used to animate a different target identity is \emph{cross-reenactment}.
In both cases, the appearance of the synthesized video is derived from the target identity.
This terminology allows us to formally state our goal:
we want to verify that a target video is a self-reenactment.
With this terminology in mind, we introduce our dataset, which includes real videos as well as self- and cross-reenactment videos.
\vspace{-0.03in}
\begin{figure*}[t!]
    \centering\includegraphics[width=\textwidth]{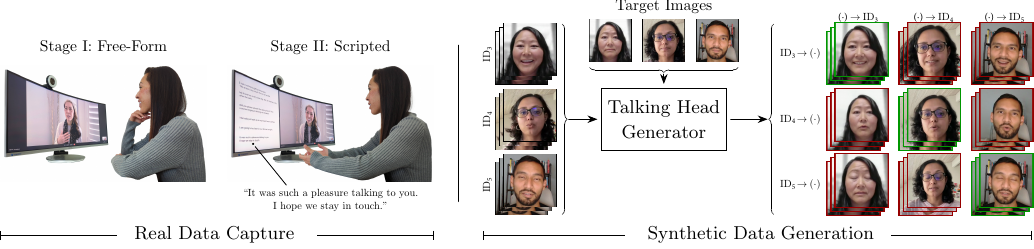}
    \vspace{-0.18in}
    \caption{We introduce the NVFAIR dataset, containing real and synthetic talking-head videos.
    We capture subjects talking in both scripted and free-form settings.
    To encourage natural performance, we record the subjects while videoconferencing with each other (left).
    We then synthesize more than $650,000$ talking-head videos---the largest collection till date---using three state-of-the-art face-reenactment talking-head generators.
    On the right, each row corresponds to a driving identity ($\scriptstyle\text{ID}_i\rightarrow(\cdot)$) and each column corresponds to a different target identity ($\scriptstyle(\cdot)\rightarrow\text{ID}_i$).
    The videos in which driving and target identity match are \textcolor{darkgreen}{self-reenactments}, the rest are \textcolor{darkred}{cross-reenactments}.}\label{fig:dataset} 
    \vspace{-0.1in}
\end{figure*}

\section{The NVIDIA Facial Reenactment (NVFAIR) Dataset}\label{sec:dataset}
\vspace{-0.05in}
Recall that avatar fingerprinting is not about detection of synthetic media.
Rather, we already know a video to be synthesized, and seek to verify that the driving identity is authorized.
This new task dictates a set of requirements for the dataset to be effective for training and evaluation.
Specifically, we need a dataset that contains
\begin{enumerate}[noitemsep,topsep=0pt,parsep=0pt,partopsep=0pt]
    \item multiple real videos per identity, with scripted and free-form conversations, and with both natural and prescribed emotions,
    \item self- \emph{and} cross-reenactments of target identities, with cross-reenactments driven by all subjects to allow for a variety of driving facial structures, and
    \item multiple face-reenactment generators.
\end{enumerate}
All relevant existing datasets only capture a subset of these requirements (see Table~\ref{table:dataset_features}; the Supplement contains further discussion).
Moreover they do not use the state-of-the-art talking-head generation technology to synthesize the self- and cross-reenactment videos.
We introduce the NVFAIR dataset that features \emph{all} of the above properties.
With over $650,000$ synthetic videos, it provides $10\times$ as many videos as the next largest dataset, and uses three different state-of-the-art generators for the self- and cross-reenactment videos.
Figure~\ref{fig:dataset} shows an overview of data capture and synthesis.

\begin{table*}
    \setlength\tabcolsep{2pt}
    \renewcommand{\arraystretch}{1.25}
    \centering
    \scriptsize{
        \begin{tabular}{r|c|c|c|c|c|c|c|c}
        Dataset          & \makecell{\# Subjects \\ (source)} & \makecell{(F)ree / \\(S)cripted?} & \makecell{Emotion:\\ (N)atural /\\(P)rescribed}& \makecell{(R)eal / \\(S)elf- /\\(C)ross-reenact.?} & \makecell{\#Face\\Reenact.}& \makecell{Avg. Videos\\per Subject} & \makecell{\# Face-\\reenact. \\Generators} & \makecell{Ethnic\\Diversity}\\
        \hline
        \tiny{RAVDESS}~\cite{livingstone2018ryerson}  &    24 (new)    &        (S)              &     (P)                   & (R) only & N/A & 120 (R) & N/A & \textcolor{darkgreen}{\checkmark}\\
        \tiny{MEAD}~\cite{wang2020mead}  &    60 (new)    &        (S)             &     (P)                  & (R) only & N/A & 720 (R) & N/A & \textcolor{darkgreen}{\checkmark}\\        
        \tiny{CREMA-D}~\cite{cao2014crema}  &      91 (new) &        (S)              &     (P)                   & (R) only &  N/A & 81 (R) & N/A & \textcolor{darkgreen}{\checkmark}\\
        \tiny{VFHQ}~\cite{fox2021}  &    36 (\cite{livingstone2018ryerson})  &        (S)             &     (P)+(N)                 & (R)+(S) & 1,737 & 120 & 1 & \textcolor{darkgreen}{\checkmark}\\
        \tiny{FF++}~\cite{rossler2019faceforensics}    &   1000  (YT)    &      (F)              &    (N)                   & (R)+(C) & 2,000 & 1(R) + 2(C) &  2 & \textcolor{darkgreen}{\checkmark}\\       
        \tiny{KoDF}\cite{kwon2021kodf}& \textbf{403 (new)} & \textbf{(F)+(S)} & (N) & (R)+(C)& 61,000 & 150(R) + 151(C) & 1 & \textcolor{darkred}{\ding{55}}\\
        \hline
        \makecell{\textbf{NVFAIR}\\\textbf{(Ours)}}    &    \makecell{161 (46 new \\+ \cite{livingstone2018ryerson}, \cite{cao2014crema})}     &        \textbf{(F)+(S)} &     \textbf{(P)+(N)}  & \textbf{(R)+(S)+(C)} &  \textbf{654,726} &\rule{0pt}{4ex} \makecell{\textbf{76(R)  + 228(S)}\\+ \textbf{3,840(C)}} & \textbf{3} & \textcolor{darkgreen}{\checkmark}\\
        \end{tabular}
    }
    \vspace{0.08in}
    \caption{The existing talking-head video datasets were designed for tasks such as deepfake detection or facial emotion analysis.
    Avatar fingerprinting is a fundamentally different task. 
    As a result, no existing dataset satisfies the requirements for training and evaluating models for avatar fingerprinting.
    To overcome this limitation, we introduce the NVFAIR dataset, which is the first dataset that offers the complete set of monologue modalities, and features the largest collection of facial reenactments to date.
    Specifically, it provides scripted and free-form monologues, with natural and prescribed emotions, and self- and cross-reenactments (driven by \emph{all} remaining subjects) synthesized using three generators, alongside original videos for newly-recorded subjects.
    }\label{table:dataset_features}
    \vspace{-0.25in}
\end{table*}


\subsection{Real Data Capture}\label{sec:data_capture}
Capturing videos of monologues delivered by different subjects for the purpose of identity verification introduces two conflicting goals.
On the one hand a controlled evaluation of the trained models requires predictability of what is spoken to prevent identification algorithms from latching onto the spoken content itself.
On the other, we want the subjects to act as they would in a casual conversation, rather than reciting memorized text, to capture their uniquely identifying mannerisms.
We address this trade-off by recording the subjects while videoconferencing in pairs, which creates the impression of being in a natural conversation.
This differs from existing datasets, in which the subjects look at the camera, but are not interacting with another person during the recording~\cite{kwon2021kodf,cao2014crema,livingstone2018ryerson}.
We also design two distinct recording strategies:
a free-form stage where the subjects are given only general guidance on the topics, and a more controlled scripted stage in which subjects speak short, memorized monologues of 2-3 sentences each, see Figure~\ref{fig:dataset}(a).
To capture the variability of real-life scenarios, we provided minimal instructions on how to setup the video call, allowing for diverse face, scale, and lightning, bandwidth stability, and background scene clutter.
In total we record 46 subjects of diverse genders, ages, and ethnicities, while strictly abiding by a pre-approved IRB protocol (see Supplement for details and privacy considerations).

\paragraph{Stage I: Free-Form Monologues.}
In this first stage, the two subjects on the call alternate between asking and answering seven pre-defined questions.
The questions are designed to avoid sensitive or potentially inflammatory topics.
This is critical because we later use sentences spoken by one individual to animate the video of a second individual, quite literally putting words in their mouths.
The complete list of questions is in the Supplement.
To further create a natural interaction, the subject listening is encouraged to actively but silently engage with the one speaking (\eg, by nodding or smiling).

\paragraph{Stage II: Scripted Monologues.}
For this stage, we prepared thirty short utterances consisting of two or three sentences each.
We chose this length to allow for memorization, while still providing enough content to trigger facial expressions.
However, to avoid inducing unnatural expressions, we do not prescribe specific emotions for each utterance.
For instance, we do not ask to express anger for a sentence, but we do choose sentences that may naturally evoke it, and used punctuation to encourage it, \eg, ``Will you please answer the darn phone? The constant ringing is driving me insane!''
We instruct the subjects to split their screens to show both this list and video call and encourage them to speak to their recording partner when reciting, see Stage II in Figure~\ref{fig:dataset}(a).
More details, including the full list of utterances can be found in the Supplement.

\subsection{Synthetic Talking-Head Videos}\label{sec:synt}

Using the videos described in Section~\ref{sec:data_capture}, as well as the original videos from the CREMA-D~\cite{cao2014crema} and RAVDESS~\cite{livingstone2018ryerson} datasets, we generate synthetic talking-head videos to train and evaluate our avatar fingerprinting algorithm.
Specifically, we pool the 91 identities from the original videos of CREMA-D~\cite{cao2014crema}, the 24 identities from those of RAVDESS~\cite{livingstone2018ryerson}, and the 46 from our own video-conferencing data capture, for a total of 161 unique identities $\mathcal{I}$.
Recall that we have several real videos for each identity $\text{ID}_i \in \mathcal{I}$.
To avoid a combinatorial explosion of synthetic videos, for all pairs of identities  $\text{ID}_i$ and $\text{ID}_j$, we use $\text{ID}_j$ as the target identity and we randomly select 8 of the videos of $\text{ID}_i$ to generate 8 cross-reenactment videos, $\{\text{ID}_i^k\rightarrow\text{ID}_j\}_{k=\{1,..,8\}}$ (all 8 share the same target image).
We also generate self-reenactment videos for each of the target identities, by animating their neutral-face images derived from captured videos with each of their real videos.

We use three different generators for synthesizing the videos for all the 161 identities: face-vid2vid~\cite{wang2021facevid2vid}, LIA~\cite{wang2022latent}, and TPS~\cite{tps2022}.
This allows us to test if our model generalizes across generators. We chose these talking-head generators because they are the state of the art and they preserve the identity-specific facial motion dynamics well.
Nevertheless, the reconstruction is not perfect;
for instance, in the third row of Figure~\ref{fig:dataset}(b) the person in the driving video (ID$_5$) is squinting, but the eyes are shut in all the synthetic videos, including the self-reenactment video.
In total we generate more than $650,000$ synthetic videos, which required more than $2,500$ RTX 3090 GPU hours.
More details are in the Supplement.

\section{Method}\label{sec:method}

\paragraph{Overview.} We seek to verify the driving identity of a synthetic video, independently of the target identity.
We leverage the finding from cognitive science research that each person emotes in unique ways when communicating, and that this signal is sufficient for recognition, even when the actual appearance is artificially corrupted~\cite{OToole_2002_CogSci,Hill_2001_CurrentBio,Knappmeyer_2001_JoV}.
We note that our method does not latch onto artifacts introduced by the generators---a property that we demonstrate by showing generalization to new generators not seen during training.
Rather, our features capture the dynamics of the expressions, like the way a person frowns, or the way she smiles.
Notably, they are distinct from the temporal artifacts introduced by the generator, and that existing algorithms use to detect whether a video is synthetic or real~\cite{haliassos2021lips}. 

\begin{figure}[t]
    \centering\includegraphics[width=0.72\columnwidth]{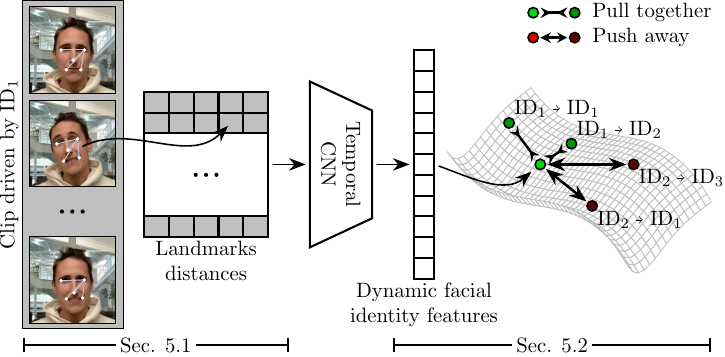}
    \vspace{-0.1in}
    \caption{We extract landmarks from the frames of a talking-head clip, compute their normalized pairwise distances, and concatenate the frame-wise features.
    We then learn an identity embedding using a loss that pulls closer features of videos driven by the same identity and pushes away those driven by others.
    $\text{ID}_i\rightarrow\text{ID}_j$ indicates a video that looks like identity $j$ (the ``target'' identity), and is driven by identity $i$.}\label{fig:method}
    \vspace{-0.2in}
\end{figure}

An overview of our algorithm for avatar fingerprinting is shown in Figure~\ref{fig:method}.
To capture expressions, we extract the relative position of facial landmarks over time from the input video, as shown in Figure~\ref{fig:method} (Section~\ref{sec:features}).
We learn to project these temporal signatures onto a \emph{dynamic identity embedding} in which features belonging to the same driving identity are close to each other regardless of the target identity, \ie, independently of appearance (Figure~\ref{fig:teaser}).
To learn this embedding we train a 3DCNN (where the third dimension spans video frames---a \textit{temporal} CNN) with a novel contrastive loss that pulls together all embedding vectors of synthetic videos driven by an individual, while pushing away the embedding vectors of videos driven by all others individuals (Section~\ref{sec:loss}).
More implementation details are in Section~\ref{sec:implem}.

\subsection{Dynamic Facial Identity Features}\label{sec:features}
Our first step is to extract temporal features that summarize short segments of the video we wish to fingerprint.
We identify the following guiding principles for the extracted features.
We would like features that:
\begin{enumerate}[noitemsep,topsep=0pt,parsep=0pt,partopsep=0pt]
    \item have minimal dependency on the appearance of the face in the video (that is, the target identity), 
    \item reflect the dynamics of the expressions, and
    \item capture subtle expressions. 
\end{enumerate}
One choice could be per-frame 3DMM features~\cite{3dmm}: a strategy also used by Cozzolino~\etal to detect synthetic videos~\cite{Cozzolino_2021_ICCV}.
However, we empirically observe that 3DMM features are not sufficiently expressive, and do not satisfy desideratum 3 (see ablation experiments in Section~\ref{sec:results}).
We observe a similar behavior for action units~\cite{Agarwal_2019_CVPR_Workshops}.
Facial landmarks~\cite{honari2018improving} address this issue, but are sensitive to the shape of the face in the video, and thus to the target identity.

To leverage the expressiveness of facial landmarks while abstracting from the underlying facial shape, we compute the pairwise normalized Euclidean distances between each pair of landmarks of a frame.
We concatenate these distances into a single vector for the frame, $\frameFeat$.
A subset of the facial landmarks and distances are shown in Figure~\ref{fig:method}.

We then break the input video into \emph{clips}, which are sequences consisting of F frames and offset by one frame (\eg, [1,F], [2,F+1], \etc), and concatenate the vectors from all the frames in each clip.
Using the change in the relative position of the landmarks over a short period of time (the length of a clip) allows us to capture temporal dynamics with minimal dependence on the absolute position of each landmark, \ie, independently of the shape of the face.

We show empirically that our features are a good representation for our task, by comparing against alternative choices for input features such as 3DMM (Section~\ref{sec:results}).

\subsection{Dynamic Identity Embedding Contrastive Loss}\label{sec:loss}
While the features described in Section~\ref{sec:features} are designed to extract low-level motion dynamics, they cannot be used directly to disambiguate two target videos based on their driving identities.
We tackle this problem by learning a dynamic identity embedding, a space where videos driven by one subject map to points that are close to each other and far from the videos driven by anybody else.

Specifically, we use a temporal 3DCNN to extract an embedding vector from a clip, which, as described before, is a short segment of an input video.
To train the network we use a dataset of synthetic videos driven by different identities.
We denote as $\clip{1}{2}^{k}(t)$ the embedding produced by the network for the clip starting at time $t$ in the $k$-th video, of a target identity ID$_2$ driven by identity ID$_1$.
As stated above, we have two main objectives, which we capture with the following terms in our proposed loss function.

\paragraph{We Want to Pull Together All the Videos Driven by ID$_1$.}
To achieve this, we define the following term:
\begin{equation}
    \text{N}_{j,\text{ID}_1,\text{ID}_2}(t) = \sum_{\text{ID}_l, k} \max_n s( \clip{1}{2}^{j}(t), \clip{1}{l}^{k}(n)),
    \label{eq:pull}
\end{equation}
where $s(\cdot ,\cdot) = e^{- \rVert \cdot, \cdot \lVert^2}$ is a similarity metric.
Intuitively, Equation~\ref{eq:pull} takes two videos, $j$ and $k$, both driven by ID$_1$.
Given a clip starting at time $t$ in the first video, it looks for the most similar clip in the second video.
Since the driving identity is the same for both videos, Equation~\ref{eq:pull} encourages an embedding where clips that capture a similar expression are closer to each other.
Equation~\ref{eq:pull} is high even if only one clip from video $k$ has a similar temporal signature to $\clip{1}{2}^{j}(t)$.
That is because even just one occurrence of the same expression is evidence that the driving identity may be the same.
Of course, other driving identities may use similar expressions and we address that with the loss term described below.
Additionally, we note that $k$ spans the set of \emph{all} videos driven by ID$_1$, and ID$_l$ spans \emph{all} identities, including $\text{ID}_l=\text{ID}_1$ and $\text{ID}_l=\text{ID}_2$.

\paragraph{We Want to Push Away Videos not Driven by ID$_1$.}
We define the following term:
\begin{equation}
    \text{Q}_{j,\text{ID}_1,\text{ID}_2}(t) = \sum_{\text{ID}_l \neq \text{ID}_1,k} \max_n s( \clip{1}{2}^{j}(t) , \clip{l}{2}^{k}(n)),
    \label{eq:push}
\end{equation}
where, similarly to Equation~\ref{eq:pull}, we take a clip from video $j$, and look for the most similar clip in video $k$.
This time the two videos share the same target identity, but are driven by different identities:
we want all the videos driven by identities different from ID$_1$ to be pushed away from those driven by ID$_1$, including videos where ID$_1$ is the target identity.
Note that ID$_2$ spans \emph{all} identities, including $\text{ID}_2=\text{ID}_1$ and $\text{ID}_2=\text{ID}_l$.

\paragraph{We Want to Rely on the \textit{Temporal} Dynamics of the Videos Driven by ID$_1$.}
Although we use a temporal CNN, the model could still learn to rely on static expressions, such as a snapshot of the person smiling, rather than the \textit{temporal} progression of expression leading to, or following, the smile.
To further encourage the model to learn from the temporal dynamics, we introduce an additional term:
\begin{equation}
    \text{R}_{j,\text{ID}_1,\text{ID}_2}(t) = \sum_{\text{ID}_l, k} \max_n s( \clip{1}{2}^{j}(t), \shuffledclip{1}{l}^{k}(n)),
    \label{eq:timeshuffle-push}
    \vspace{-0.1in}
\end{equation}
where $\shuffledclip{1}{l}^{k}$ denotes a version of the clip $\clip{1}{l}^{k}$ from Equation~\ref{eq:pull} with randomly shuffled frame ordering.
We want such time-shuffled versions of the clips driven by ID$_{1}$ to be pushed away from the pristine clips driven of ID$_{1}$. 
Effectively, this means that the driving identity of the time-shuffled clips is regarded as different from ID$_{1}$.
In other words, we want to pull together video clips in the learned embedding space only when the temporal facial dynamics are characteristic of ID$_{1}$.
We further show the importance of this term in Section~\ref{sec:ablation}.

Combining Equations~\ref{eq:pull},~\ref{eq:push}, and~\ref{eq:timeshuffle-push}, we write the probability that the embedding vector $\clip{1}{2}^{j}(t)$ lies close to the embedding vectors for \emph{all} video clips driven by ID$_1$ and far from \emph{all} the videos driven by others as
\begin{equation}
    p_{j,\text{ID}_1,\text{ID}_2}(t) = \frac{\text{N}_{j,\text{ID}_1,\text{ID}_2}(t)}{\text{N}_{j,\text{ID}_1,\text{ID}_2}(t) + \text{Q}_{j,\text{ID}_1,\text{ID}_2}(t)+\text{R}_{j,\text{ID}_1,\text{ID}_2}(t)},
\end{equation}
and the complete loss term as
\begin{equation}
    \mathcal{L} = \sum_{j,\text{ID}_1,\text{ID}_2,t} -\text{log}(p_{j,\text{ID}_1,\text{ID}_2}(t)).
    \label{eq:loss}
    \vspace{-0.in}
\end{equation}


\subsection{Implementation}\label{sec:implem}
\paragraph{Parameter Choices.}
To extract the per-frame dynamic facial identity features $\frameFeat$, we detect 126 facial landmarks for each frame~\cite{honari2018improving}, and compute the per-frame normalized pairwise Euclidean distances between these landmarks.
The clip duration is set to 71 frames.
We find that this is sufficient to capture the facial dynamics that are meaningful for avatar fingerprinting, while also maintaining a good trade-off between speed and accuracy.
We also experiment with shorter-duration video clips (see Supplement).
The input tensor to the temporal CNN is obtained by concatenating $\frameFeat$ across 71 frames.
In each batch, we include 8 unique identities.
For each identity $\text{ID}_i$, the pull term (Equation~\ref{eq:pull}) comprises 16 clips: 8 are self-reenactments, randomly sampled from the full set, and the remaining are cross-reenactments with $\text{ID}_i$ as the \emph{driving} identity.
This allows the neural network to pull together videos based purely on the facial motion, regardless of the appearance of the video.
The push term (Equation~\ref{eq:push}) for $\text{ID}_i$ consists of clips with the remaining 7 identities in the batch serving as driving identities (8 clips per driving identity), as well as the time-shuffled self-reenactments of $\text{ID}_i$ (Equation~\ref{eq:timeshuffle-push}).
Additional training details can be found in the Supplement.

\vspace{-0.1in}
\paragraph{Training, Validation, and Testing Datasets.}
Of the 161 total identities (pooling together the identities from our dataset, RAVDESS, and CREMA-D, see Section~\ref{sec:synt}), we reserve 35 for testing, 14 for validation, and 112 for training.
We ensure that there are no cross-set cross-reenactments: that is, identities in the training set only drive other training-set identities (and similarly for the validation and test sets).
This allows us to test the models on facial appearances and expressions that were not seen during training.
To evaluate the generalization of trained models to new generators, we train our network on videos generated for the training set identities using one generator, and evaluate on the synthetic videos of test-set identities synthesized using remaining two generators.

\section{Evaluation}\label{sec:results}
We thoroughly evaluate our algorithm both qualitatively and quantitatively.
Our algorithm outperforms reasonable baselines (Section~\ref{sec:baselines}), it generalizes to generators not seen in training (Section~\ref{sec:generalization}), and it is robust to video compression (Supplement).
We also perform a number of ablation studies to analyze our design choices: our input features, the importance of the time-shuffling term in the loss function (Section~\ref{sec:ablation}), and the impact of clip duration (Supplement).

\begin{figure}[t!]
    \centering
    \input{figures/results/videoresults}
\end{figure}

We begin by evaluating qualitatively our method's ability to extract embedding vectors based on the driving identity.
Figure~\ref{fig:videos} shows a set of self- and cross-reenacted clips (please view the animation in a media-enabled viewer, such as Adobe Acrobat).
For each row, we take one identity as reference and we compute the embedding vectors of clips that use it both as the driving and the target identity.
We then compute the average Euclidean distance of the resulting embedding vectors against those of \emph{the self-reenactments by the same reference identity}.
We note that the average distance $d$ is lower when the driving identity matches the reference identity (first two columns).
We also note that the distances between the clips in the first two columns are similar:
this confirms that the distance is a function of the facial motion, rather than the facial appearance.
When the driving identity changes, the average distance increases, even if the target identity matches the reference identity, which is precisely our goal.
More results are in the Supplement.

To evaluate our approach more formally, we use the 35 unique test-set identities that are not used as driving or target identities in the training set (Section~\ref{sec:implem}).
One at a time, we treat each identity ID$_i$ as target and synthesize cross-reenactments using all the remaining identities as drivers.
This is the set of ``unauthorized'' synthetic videos for ID$_i$.
The self-reenacted samples for ID$_i$ form the ``authorized'' set.
Note that there are several self-reenacted videos of ID$_i$, one per original video of ID$_i$.

For each target identity ID$_i$, we extract the dynamic identity embedding vector of all the clips in the pool of its self- and cross-reenacted videos, and compute their Euclidean distances.
That is, for clip $k$ we compute 
\begin{equation}
    \begin{split}
    &d(\clip{i}{i}^k, \clip{i}{i}^l),~~\forall l \neq k,~~\text{and}\\
    &d(\clip{i}{i}^k, \clip{j}{i}^l),~~\forall l \neq k,~~~\forall i \neq j.
    \end{split}
\end{equation}
We threshold these distances for each target identity to get an ROC curve, and average across the target identities to get the overall area under the curve (AUC).
We note that this AUC measures one model's ability to classify a synthetic video as self-reenactment or as cross-reenacted.
We conduct further analysis of our model's ability to classify other categories of videos---such as, evaluating AUC on same-utterance self- vs. cross-reenactments or on scripted vs. free-form monologues---in the Supplement.

\begin{figure}[t!]
    \centering
        \resizebox{0.3\columnwidth}{!}{%
            \begin{tikzpicture}
                \hspace{-3in}
                \begin{axis}[name=plot1,xmin=0,xmax=1,ymin=0,ymax=1,xtick={0,0.5,1},ytick={0,0.5,1},legend style={legend pos=south east,font=\Large}]
                    \addplot[mark='',color=ROC1, line width=2pt]  table [x=fpr, y=tpr, col sep=comma] {figures/baseline_eval/liveportait_ourLoss_test_ROC_liveportrait_52799_presaved_normfac.csv};
                    \addplot[mark='',color=ROC3, line width=2pt] table [x=fpr, y=tpr, col sep=comma] {figures/baseline_eval/Agarwal_test_ROC_liveportrait.csv};
                    \addplot[mark='',color=ROC2, line width=2pt] table [x=fpr, y=tpr, col sep=comma] {figures/baseline_eval/IDreveal_test_ROC_liveportrait.csv};
                    \legend{\textbf{Ours} -- 0.87, \cite{Agarwal_2019_CVPR_Workshops} -- 0.73, \cite{Cozzolino_2021_ICCV} -- 0.72}
                \end{axis}
                \node[above=2pt of plot1.north]{\LARGE Generator: Face-vid2vid};
                \hspace{3in}
                \begin{axis}[name=plot2,xmin=0,xmax=1,ymin=0,ymax=1,xtick={0,0.5,1},ytick={0,0.5,1},legend style={legend pos=south east,font=\Large}]
                    \addplot[mark='',color=ROC1, line width=2pt]  table [x=fpr, y=tpr, col sep=comma] {figures/baseline_eval/tps_ourLoss_test_ROC_tps_75299_presaved_normfac.csv};
                    \addplot[mark='',color=ROC3, line width=2pt] table [x=fpr, y=tpr, col sep=comma] {figures/baseline_eval/Agarwal_test_ROC_tps.csv};
                    \addplot[mark='',color=ROC2, line width=2pt] table [x=fpr, y=tpr, col sep=comma] {figures/baseline_eval/IDreveal_test_ROC_tps.csv};
                    \legend{\textbf{Ours} -- 0.85, \cite{Agarwal_2019_CVPR_Workshops} -- 0.68, \cite{Cozzolino_2021_ICCV} -- 0.71}
                \end{axis}      
                \node[above=2pt of plot2.north]{\LARGE Generator: TPS};
                \hspace{3in}
                \begin{axis}[name=plot3,xmin=0,xmax=1,ymin=0,ymax=1,xtick={0,0.5,1},ytick={0,0.5,1},legend style={legend pos=south east,font=\Large}]
                    \addplot[mark='',color=ROC1, line width=2pt]  table [x=fpr, y=tpr, col sep=comma] {figures/baseline_eval/lia_ourLoss_test_ROC_lia_50399_presaved_normfac.csv};
                    \addplot[mark='',color=ROC3, line width=2pt] table [x=fpr, y=tpr, col sep=comma] {figures/baseline_eval/Agarwal_test_ROC_lia.csv};
                    \addplot[mark='',color=ROC2, line width=2pt] table [x=fpr, y=tpr, col sep=comma] {figures/baseline_eval/IDreveal_test_ROC_lia.csv};
                    \legend{\textbf{Ours} -- 0.84, \cite{Agarwal_2019_CVPR_Workshops} -- 0.67, \cite{Cozzolino_2021_ICCV} -- 0.67}
                \end{axis}
                \node[above=2pt of plot3.north]{\LARGE Generator: LIA};
            \end{tikzpicture}
    }
    \vspace{-0.1in}
    \caption{ROC curves and AUC values for our method and two baselines: Agarwal~\etal~\cite{Agarwal_2019_CVPR_Workshops} and ID-Reveal~\cite{Cozzolino_2021_ICCV}. Each sub-plot shows the results on our test set for each of the three talking-head generators: face-vid2vid~\cite{wang2021facevid2vid}, LIA~\cite{wang2022latent}, and TPS~\cite{tps2022}.}\label{fig:auc}
    \vspace{-0.15in}
\end{figure}

\subsection{Comparisons with Existing Methods}\label{sec:baselines}
Avatar fingerprinting is a novel task, and no existing methods directly address it.
The closest related works aim at detecting real versus synthetic media.
As discussed in Section~\ref{sec:related}, some of these detectors learn identity-specific features such as facial expressions and head poses~\cite{Agarwal_2019_CVPR_Workshops}, or facial shapes and motion~\cite{Cozzolino_2021_ICCV} and can serve as baselines for the task of avatar fingerprinting with some adaptation.
The work by Agarwal~\etal trains a model to detect synthetic videos of a \emph{specific} identity~\cite{Agarwal_2019_CVPR_Workshops}.
To adapt it to our task, we train 35 different models, one for each identity in the evaluation, by splitting the corresponding original videos into two subsets.
We then test each model on the self- and cross-reenactment videos of the corresponding identity.
ID-Reveal, trained on a large-scale dataset, learns an embedding space where real videos of a specific identity are grouped together~\cite{Cozzolino_2021_ICCV}.
Since it shows good generalization to new identities, for the task of synthetic media detection, we directly use the pre-trained model on our data to detect, once again, self- versus cross-reenactment.
Figure~\ref{fig:auc} shows the ROC curves for our method compared to these baselines, on three face-reenactment generators (face-vid2vid~\cite{wang2021facevid2vid}, LIA~\cite{wang2022latent}, and TPS~\cite{tps2022}).
Our method (AUC=0.868 on face-vid2vid) outperforms by a wide margin both ID-Reveal (AUC=0.720 on face-vid2vid), and the method by Agarwal~\etal (AUC=0.726 on face-vid2vid).
We also note that, unlike ID-Reveal and our method, Agarwal~\etal uses a different model per identity.

\begin{figure}[t!]
    \centering
    \resizebox{0.3\columnwidth}{!}{%
    \begin{tikzpicture}
        \hspace{-3in}
        \begin{axis}[name=plot1,xmin=0,xmax=1,ymin=0,ymax=1,xtick={0,0.5,1},ytick={0,0.5,1},legend style={legend pos=south east,font=\Large}]
            \addplot[mark='',color=ROC1, line width=2pt]  table [x=fpr, y=tpr, col sep=comma] {figures/ablations/liveportait_ourLoss_test_ROC_liveportrait_52799_presaved_normfac.csv};
            \addplot[mark='',color=ROC3, line width=2pt] table [x=fpr, y=tpr, col sep=comma] {figures/ablations/liveportait_ourLoss_test_ROC_tps_52799_presaved_normfac.csv};
            \addplot[mark='',color=ROC2, line width=2pt] table [x=fpr, y=tpr, col sep=comma] {figures/ablations/liveportait_ourLoss_test_ROC_lia_52799_presaved_normfac.csv};
            \legend{Face-vid2vid -- 0.87, TPS -- 0.82, LIA -- 0.84}
        \end{axis}
        \node[above=2pt of plot1.north]{\LARGE Trained on Face-vid2vid~\cite{wang2021facevid2vid}};
        \hspace{3in}
        \begin{axis}[name=plot2,xmin=0,xmax=1,ymin=0,ymax=1,xtick={0,0.5,1},ytick={0,0.5,1},legend style={legend pos=south east,font=\Large}]
            \addplot[mark='',color=ROC1, line width=2pt]  table [x=fpr, y=tpr, col sep=comma] {figures/ablations/tps_ourLoss_test_ROC_liveportrait_75299_presaved_normfac.csv};
            \addplot[mark='',color=ROC3, line width=2pt] table [x=fpr, y=tpr, col sep=comma] {figures/ablations/tps_ourLoss_test_ROC_tps_75299_presaved_normfac.csv};
            \addplot[mark='',color=ROC2, line width=2pt] table [x=fpr, y=tpr, col sep=comma] {figures/ablations/tps_ourLoss_test_ROC_lia_75299_presaved_normfac.csv};
            \legend{Face-vid2vid -- 0.85, TPS -- 0.85, LIA -- 0.84}
        \end{axis}
        \node[above=2pt of plot2.north]{\LARGE Trained on TPS~\cite{tps2022}};
        \hspace{3in}
        \begin{axis}[name=plot3,xmin=0,xmax=1,ymin=0,ymax=1,xtick={0,0.5,1},ytick={0,0.5,1},legend style={legend pos=south east,font=\Large}]
            \addplot[mark='',color=ROC1, line width=2pt]  table [x=fpr, y=tpr, col sep=comma] {figures/ablations/lia_ourLoss_test_ROC_liveportrait_50399_presaved_normfac.csv};
            \addplot[mark='',color=ROC3, line width=2pt] table [x=fpr, y=tpr, col sep=comma] {figures/ablations/lia_ourLoss_test_ROC_tps_50399_presaved_normfac.csv};
            \addplot[mark='',color=ROC2, line width=2pt] table [x=fpr, y=tpr, col sep=comma] {figures/ablations/lia_ourLoss_test_ROC_lia_50399_presaved_normfac.csv};
            \legend{Face-vid2vid -- 0.83, TPS -- 0.82, LIA -- 0.84}
        \end{axis}
        \node[above=2pt of plot3.north]{\LARGE Trained on LIA~\cite{wang2022latent}};
    \end{tikzpicture}
}
    \vspace{-0.1in}
    \caption{\textbf{Generalization to new generators}. To study the robustness to new talking-head generators, we train three version of our model on three different generators and test on all three.}
    \label{fig:new-generator-robust}
    \vspace{-0.15in}
\end{figure}

\subsection{Generalization to New Generators}\label{sec:generalization}
For an avatar fingerprinting model algorithm to be broadly applicable, generalization to new talking-head generators that are not seen in training is crucial.
Since our dataset contains videos synthesized by three different generators, we can train three models, one with each generator, and test these models on all three generators. 
Figure~\ref{fig:new-generator-robust} shows the resulting ROC curves and AUC values: the overlap of the curves and similar AUC values in each subplot confirms that our method generalizes well to new generators.

\subsection{Ablation Study}\label{sec:ablation}
Our method outperforms existing baselines by introducing two novel components:
the dynamic facial identity features, which capture the facial dynamics in a compact and expressive way, and the loss function, which defines the shape of the identity embedding.
Here we study the contribution of each, using the face-vid2vid generator for training and testing.
We evaluate the contribution of our dynamic facial identity features by swapping them with 3DMM features~\cite{3dmm1999}, a popular choice to capture facial dynamics.
Since we use a temporal CNN backbone similar to the one from ID-Reveal, for this ablation we use the loss function proposed in their original paper~\cite{Cozzolino_2021_ICCV}.
We re-train the same network using our features and observe a jump from 0.718 to 0.754 in terms of AUC.
Upon inspection we notice that the 3DMM features tend to over-smooth the facial motion, and are unable to capture subtle dynamics that prove critical to avatar fingerprinting, and which our features capture.
We also evaluate the contribution of our dynamic identity embedding loss and observe a further improvement (AUC 0.868).
With our loss formulation, the advantage of $\text{R}_{j,\text{ID}_1,\text{ID}_2}(t)$ in Eq.~\ref{eq:loss} is also evident when compared against a model trained without this term (AUC 0.850).
Table~\ref{table:feature-loss-ablation} summarizes this ablation study.
Additional experiments in the Supplement show the impact of $F$, performance on scripted vs. free-from monologues, and robustness to video distortions.

\begin{table}[t!]
    \setlength\tabcolsep{5pt}
    \renewcommand{\arraystretch}{1.25}
    \centering
    \resizebox{0.65\columnwidth}{!}{
    \scriptsize{
        \begin{tabular}{c|c|c}
        Input Features    & Loss                  &       AUC     \\\hline
        3DMM            & ID-Reveal rec. loss~\cite{Cozzolino_2021_ICCV}    &        0.718     \\
        Landmark distances         & ID-Reveal rec. loss~\cite{Cozzolino_2021_ICCV}     &        0.754     \\
        Landmark distances         & Our loss without   $\text{R}_{j,\text{ID}_1,\text{ID}_2}(t)$    &    0.850      \\
        Landmark distances         & Our loss with    $\text{R}_{j,\text{ID}_1,\text{ID}_2}(t)$      &        0.868     \\
        \end{tabular}
    }}
    \vspace{0.1in}
    \caption{Ablation study showing the importance of our input features and loss function design.}\label{table:feature-loss-ablation}
    \vspace{-0.26in}
\end{table}

\subsection{Limitations}\label{sec:limitation}

Our algorithm is less discriminative of subjects that are less emotive and more neutral.
In the future, relying on more granular dynamic signatures that can extract micro-expressions can help alleviate this.
The performance of our method degrades when expressions that are critical to verifying the driving identity are not captured by the synthetic portrait generator. 
Lastly, our dataset currently features only one style of interaction: one-on-one conversations.
We plan to expand to other conversation styles, such as one-way speeches, in future.

\section{Societal Impact}\label{sec:impact}
We acknowledge the societal importance of introducing guardrails when it comes to the use of talking-head generation technology. We present this work as a step towards trustworthy use of such technologies. Nevertheless, our work could be misconstrued as having solved the problem and inadvertently accelerate the unhindered adoption of synthetic talking-head technology. We do not advocate for this. Instead, we stress that this is the first work on this topic and underscore the importance of further research.
\vspace{-0.1in}

\section{Conclusions}\label{sec:conclusions}

Highly photo-real synthetic talking-head generators are becoming increasingly beneficial to applications such as video conferencing and AR/VR-based remote interactions.
This trend raises the important new research question of how best to also ensure their safe use in such scenarios.
To this end, we investigate the new problem of avatar fingerprinting, to authenticate legitimate talking-heads created by authorized users.
We leverage the fact that driving individuals have uniquely identifying dynamic motion signatures, which are also preserved in the synthetic videos that they drive.
Since none exists, we contribute a new large-scale dataset carefully designed for avatar fingerprinting and related tasks.
We hope that our work lays the foundation for further research.

\noindent \paragraph{Acknowledgements.} We would like to thank the data-capture participants, and Desiree Luong, Woody Luong, and Josh Holland for their help with Figure~\ref{fig:dataset}. 
We acknowledge David Taubenheim for the voiceover in the demo video, and Abhishek Badki for help with the training infrastructure.
We thank Joohwan Kim, Rachel Brown, Anjul Patney, Ben Boudaoud, Josef Spjut, Saori Kaji, Nikki Pope, and Kai Pong for their help with putting together the data capture protocols, informed consent form, photo release form, and agreements for data governance and third-party data sharing.
Koki Nagano, Ekta Prashnani, and David Luebke were partially supported by DARPA’s Semantic Forensics (SemaFor) contract (HR0011-20-3-0005). 
This research was funded, in part, by DARPA’s Semantic Forensics (SemaFor) contract HR0011-20-3-0005. The views and conclusions contained in this document are those of the authors and should not be interpreted as representing the official policies, either expressed or implied, of the U.S. Government.
\\
Distribution Statement “A” (Approved for Public Release, Distribution Unlimited).

\bibliographystyle{splncs04}
\bibliography{avatar_fingerprinting_arXiv}

\newpage
\clearpage
\renewcommand{\thesection}{A\arabic{section}}
\renewcommand{\thefigure}{A\arabic{figure}}
\renewcommand{\thetable}{A\arabic{table}}

\newpage
\null
\begin{center}
  {\Large \bf {A}vatar {F}ingerprinting for {A}uthorized {U}se of \\{S}ynthetic {T}alking-{H}ead {V}ideos\\({S}upplementary)}
  {
  \large
  \lineskip .5em
  \begin{tabular}[t]{c}
      
  \end{tabular}
  \par
  }
  \vskip .5em
  \vspace*{0pt}
\end{center}

\section{Future Work}\label{sec:supp_future_work}

In this work, we present the novel task of avatar fingerprinting, that aims to verify the authorized use of synthetic talking-head avatars, along with the first method to solve for this task as well as a novel dataset that caters precisely to the needs of avatar fingerprinting.
While talking-head avatars are becoming ubiquitous in across several applications, the avatar generation technology itself is also rapidly advancing towards generating full-body and 3D avatars~\cite{remelli2022drivable, petrovich2024multi, chan2022efficient, yuan2023gavatar}, generating realistic synthetic audio~\cite{kovela2023any}, controlling talking-head avatars with audio only~\cite{tian2024emo}, as well as controlling upper-body avatars including hands with audio only~\cite{corona2024vlogger}.
This makes our proposed first step towards verifying authorized use of synthetic talking-head avatars all the more pertinent, and paves the way for future research directions that verify the ``dynamic identity signatures'' from newer forms of avatars ranging from full-body avatars, to 3D, to audio-driven ones.

\section{Dataset}\label{sec:supp_dataset}

We now provide additional details of our proposed dataset, including details of the question prompts and sentences spoken in both stages, instructions to the participants, demographics of the dataset, and other relevant statistics.
\vspace{-0.1in}
\paragraph{General Instructions to Subjects.}
The subjects were asked to join pre-assigned Google Meet video calls using a laptop or a desktop.
For the recorded video call, the subjects were also asked to position themselves so that their face was centered and parallel to the screen.
However, in some cases with specific video-conferencing setups, this constraint was only approximately satisfied.
Additionally, subjects were instructed to avoid hand motion since it can occlude their face, and also excessive body motion that might impair the visibility of their face.
Before beginning each monologue, subjects were asked to speak ``start topic'' in a loud, clear voice, and, similarly, the end of each monologue was marked by the subjects speaking ``end topic''.
These keywords allowed for effectively using the time-stamped transcription to automatically isolate relevant portions of the Google Meet recordings.
Right after a subject said ``start topic'', they were instructed to pause for a few seconds and look directly at the camera with a frontal head pose, while holding a neutral expression. 
These frames with neutral expressions are crucial for a successful generation of synthetic talking-head videos using face-vid2vid~\cite{wang2021facevid2vid}, LIA~\cite{wang2022latent}, and TPS~\cite{tps2022}. 
These generators work by transferring expression changes from a driving video to the target image.
Therefore, it is important that the expression of the target image and that of the first frame of the driving video match.
Asking subjects to provide a neutral expression before commencing with their monologues proves to be an effective way to achieve this: these neutral frames serve as good target images, while driving videos that start with these neutral frames allow for effectively animating the target image showing a similar expression.
During the second stage of the data capture, where we record scripted monologues, subjects were instructed to memorize and speak the sentences to their recording partner, without referring back to the printed text from which they memorized the sentences.
In case the subject forgot a sentence, they were instructed to start from the beginning of the sentence set.
The whole recording session with both subjects in a video call typically lasted an hour, which also included miscellaneous interactions in between the monologues.
The current dataset release excludes such interactions and only focuses on data captured for the two stages (Free-Form Monologues and Scripted Monologues).

\vspace{-0.1in}
\paragraph{Stage I: Free-Form Monologues.}
Subjects were asked to alternate between speaker and prompter roles.
The prompter's task was to ask each of the following questions to the speaker, and the speaker was instructed to answer these questions in their natural manner.
\begin{enumerate}[noitemsep,topsep=0pt,parsep=0pt,partopsep=0pt]
    \item Describe a day when you had to rush to an appointment.
    \item Talk about an important milestone you have missed in the past and your feelings about it.
    \item What is your favorite family holiday?
    \item How is the weather in your area typically?
    \item Is there a household chore you don’t like doing?
    \item Tell me about an incident that really surprised you.
    \item Tell me about an incident that really scared you.
\end{enumerate}

\vspace{-0.1in}
\paragraph{Stage II: Scripted Monologues.} 
The following sentence sets were memorized and recited by each subject (alternating with their recording partner) in the second stage of the data capture.
We did not ask subjects to explicitly demonstrate specific emotions for any sentence set. 
Rather, we chose to allow subjects to perform these memorized sentences in a manner natural to them.

\begin{enumerate}[noitemsep,topsep=0pt,parsep=0pt,partopsep=0pt]
    \item My friend has a very cute dog. But, he can be scary when he barks.
    \item Will you please answer the darn phone? The constant ringing is driving me insane!
    \item My aunt was in the hospital for a week.
    Unfortunately, she passed away yesterday and I will need some time to grieve.    
    \item I hate rushing to get to the airport.
    The stress is too much for me to handle.     
    \item A slice of cake is the perfect ending to a meal.
    Wouldn’t you agree?    
    \item It is going to be great working with you!
    I am surprised we didn’t connect sooner!    
    \item You need to take the trash out right now!
    Your whole apartment smells like rotten eggs!    
    \item My internet connection is unreliable today.
    I hope it gets better before my meeting or I will have to call in!    
    \item I know the deadline is around the corner, but I just don’t have any updates yet, I’m sorry.
    \item Why can’t the banker figure out what’s going on?
    I should have got my money last night!    
    \item It’s really nice out today.
    I might go for a walk if I get off work early and the kids aren’t back from school.    
    \item There is a famous coffee shop around the corner that also serves snacks. 
    Would you like to go tonight?
    \item My dog almost got run over by a car today!
    Thank God he is safe!
    \item It is getting very cold outside. 
    I feel like having some hot chocolate. Would you like some?    
    \item I have been exercising so much lately.
    But I am not getting any stronger!    
    \item I have an old tie that I can wear to the interview.
    My grandfather gave it to me last year.
    \item I had fun last night - we had quite a few drinks.
    But I have a really bad hangover this morning and I am considering calling in sick.    
    \item Please don’t interrupt me when I am talking!
    Now I have forgotten what I wanted to tell you.    
    \item It was such a pleasure talking to you.
    I hope we stay in touch.    
    \item I can’t believe I misplaced my keys yet again!
    I have to leave for the airport right now.    
    \item Gosh! the boy jumped right off the cliff into the ocean.
    He is lucky he didn’t hit a rock.    
    \item The baby just spit up on my brand new clothes.
    I am going to be late for our dinner tonight.    
    \item The food smells disgusting but tastes delicious.
    How strange is that!    
    \item I was about to park when I saw a person with a gun.
    I kept driving and called the police right away.    
    \item I decided to take a nap during my lunch break.
    I am so glad I did! I feel very refreshed.    
    \item The food didn’t get delivered on time.
    We had to keep our guests waiting while we searched for options. 
    \item I was walking down an alley the other night.
    I had the strange feeling that someone was following me.    
    \item She twisted her ankle while ice-skating.
    It was her final performance for the season.    
    \item Who moved my boxes from this room? I need to find my shoes before I can head out.    
    \item We miss our old home in the mountains quite a bit.
    This new place just doesn’t feel as cozy.     

\end{enumerate}

\vspace{-0.1in}
\paragraph{Subject demographics.} 
Out of the total pool of subjects that volunteered data for our 2-stage data capture, 50$\%$ are female, 47.8$\%$ are male, and the remaining chose ``a gender not listed here".
Amongst different age groups, 37$\%$ of the participants are 25-34 years old, 32.6$\%$ are $35-44$ years old, 17.4$\%$ are 45-54 years old, 6.5$\%$ are 18-24 years old, and 6.5$\%$ are 55-64 years old.
In terms of race and ethnicity, $41.3\%$ are Caucasian, $47.8\%$ are Asian (including South Asian, East Asian, South-east Asian), $6.5\%$ are African, $2.2\%$ are Hispanic / Latino, $2.2\%$ are Pacific islander, and others remained unspecified. 

\vspace{-0.1in}
\paragraph{Synthetic Talking-Head Videos.} 
As mentioned briefly in the main paper, we pool together videos for the 46 identities from our own 2-stage data capture, along with videos from 24 identities of RAVDESS (scripted monologues only)~\cite{livingstone2018ryerson}, and 91 from CREMA-D (short scripted monologues only)~\cite{cao2014crema}, resulting in a total of 161 unique identities.
For each of these 161 identities, the remaining 160 are used to drive cross-reenactments, with 8 driving videos randomly selected from the total set of videos for each driving identity.
For any given target identity, we incorporate synthetic videos driven by \emph{every} remaining identity.
During training, such a large variety of cross-reenactments enable effectively learning an appearance-agnostic dynamic facial identity feature space.

\vspace{-0.1in}
\paragraph{Privacy Considerations.}
Face videos are sensitive data, since a person's face is a key identifier. 
We took on this task with care to ensure good data governance.
Our proposal for the data capture protocol was approved by an Institutional Review Board (IRB).
Our goal was to provide exhaustive and transparent information to participants about our data capture procedure, future plans with the dataset (including our intent to create synthetic data samples), and conditions under which future research would be conducted---by us and interested third parties. 
The participants were also asked to confirm whether their data can be used for research beyond avatar fingerprinting, and whether it could be shown in public disclosures.
Each file in our dataset is annotated with their responses.

\subsection{Importance of the proposed dataset}
As discussed in Section~\ref{sec:dataset} of the main paper, the existing datasets of facial talking-head videos satisfy a subset of the requirements of avatar fingerprinting.
This has motivated us to design our own dataset.
Here we elaborate further on additional existing datasets (over and above the ones already discussed in the main paper), to re-emphasize the relevance of our proposed dataset.

CelebDFv2~\cite{Celeb_DF_cvpr20}: This dataset features in-the-wild youtube videos of celebrities along with synthetic videos as well.
However, these do not contain any face reenactment generators---only FaceSwap generators---and do not contain any self-reenactments.
Furthermore, the real videos only contain free-form monologues for celebrities: to verify whether our model latches on to specific choice of words, we need scripted monologues in the dataset as well.
Overall, the lack of scripted monologues in real videos, and missing self-reenacted and face-reenactment generators makes it tough to use CelebDFv2 dataset for our work.

ForgeryNet~\cite{he2021forgerynet} is a diverse dataset that features a large collection of multiple types of synthetic generators.
However, the construction of the dataset caters more to traditional forensics tasks like deepfake detection.
Specifically, it lacks multiple cross-reenactments driven by all driving identities in the dataset, for each target identity.
Moreover, there are no self-reenactments in the dataset.
This limits the applicability of ForgeryNet to avatar fingerprinting.

MEAD~\cite{wang2020mead} contains a large collection of real videos of individuals.
Upon inspection, however, we observed that often subjects do not look directly at the camera (which is by design to capture different viewpoints of the face), and do not have a naturally-interactive facial behavior.
For effectively training a model for avatar fingerprinting, we require driving videos for synthetic video generation process to feature the natural facial dynamics for the driving identity.
A natural interactiveness in the facial behavior and voice intonation is a pre-requisite to capturing the dynamic identity signatures needed for avatar fingerprinting.
This makes MEAD less applicable to avatar fingerprinting, since capturing natural facial behavior of individuals is a pre-requisite from the set driving videos.
In contrast, in RAVDESS~\cite{livingstone2018ryerson} and CREMA-D~\cite{cao2014crema}, subjects look directly at the camera and are talking in a more natural conversational manner. 
This is also aligned with the natural interactive manner in which our own video-conferencing dataset is captured. 

Apart from the above-mentioned datasets, and the datasets already discussed in the main paper (CREMA-D~\cite{livingstone2018ryerson}, VFHQ~\cite{fox2021}, FaceForensics++~\cite{roessler2019faceforensicspp}, KoDF~\cite{kwon2021kodf}), there are many existing datasets catering specifically to deepfake detection, such as DFor~\cite{jiang2020deeperforensics1, jiang2021dfc20}, DFDC~\cite{dolhansky2020deepfake}, that share the above-mentioned limitations (lack of self-reenactments, not enough cross-reenactments, scripted videos missing, etc.).
This motivated us to design our own dataset for avatar fingerprinting.

\section{Implementation Details}\label{sec:supp_implementation}

\begin{figure}[t!]
    \centering
   \resizebox{0.7\columnwidth}{!}{
   \includegraphics{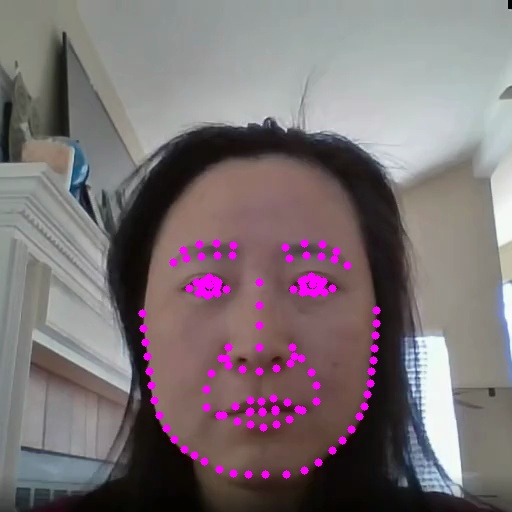}
   \includegraphics{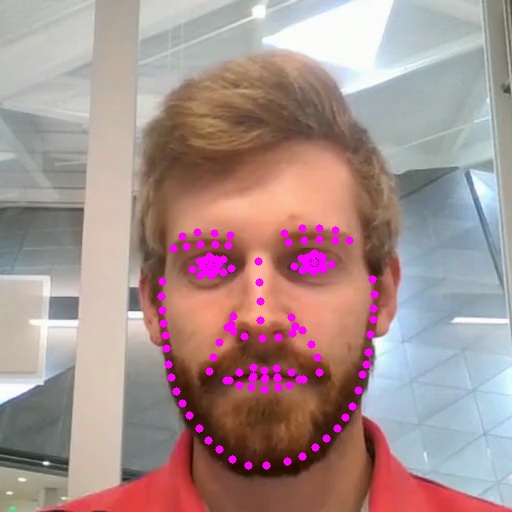}}
   \\
   \resizebox{0.7\columnwidth}{!}{
   \includegraphics{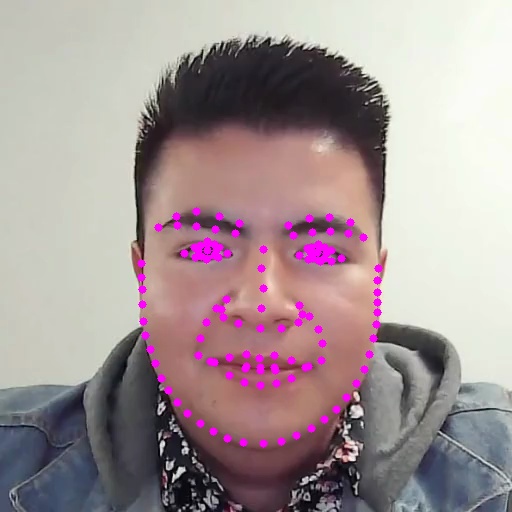}
   \includegraphics{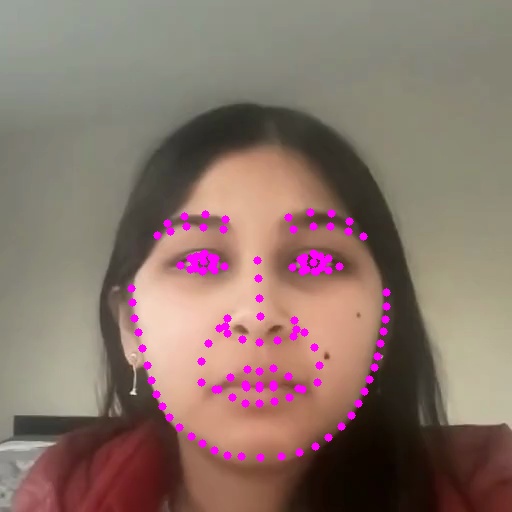}}
\vspace{-0.1in}
\caption{Visualization of all landmarks.}\label{fig:lms}

\end{figure}

Figure~\ref{fig:lms} visualizes the full set of 126 landmarks that we use to compute the input features for our model.
Our model is based on the temporal ID net~\cite{Cozzolino_2021_ICCV}.
It is trained using our input features derived from the pairwise landmark distances, after appropriately modifying the number of input channels to match our feature dimension.
To adjust the receptive field of the temporal ID net so that it predicts an embedding vector for longer or shorter input clips, we modify the number of layers of the network, and the dilation factor for the layers.
Specifically, here are the kernel sizes and dilation factors for each of the layers in the temporal ID net, depending on the choice of input  clip duration F:
\begin{enumerate}[noitemsep,topsep=0pt,parsep=0pt,partopsep=0pt]
   \item F = 31 frames: $(1, 1, 1, 1, 2, 2, 2, 2, 4)$
   \item F = 51 frames: $(1, 1, 1, 1, 2, 2, 2, 4, 4, 4, 4)$
   \item F = 71 frames: $(1, 1, 1, 1, 2, 2, 2, 2, 2, 2, 4, 4, 4, 4, 4)$
\end{enumerate}

The kernel sizes are all set to $3$ apart from the first layer, which is $1$. 
All other details of the temporal ID network are adapted from the existing implementation~\cite{Cozzolino_2021_ICCV}.
To implement the push and pull terms in Equations~\ref{eq:pull},~\ref{eq:push},~\ref{eq:timeshuffle-push} from the main paper, $n$ and $t$ span over 5 consecutive F-frame clips in a video.
That is, during training, the temporal ID net receives as input (F+4)-frame videos, and outputs 5 embedding vectors, one for each of the 5 F-frame clips in the video.
The max operation in Equations~\ref{eq:pull},~\ref{eq:push},~\ref{eq:timeshuffle-push} from the main paper is performed over the 5 different clips (therefore, 5 different values of n), and the overall loss term in Equation~\ref{eq:loss} accumulates over 5 values of $t$.
So, when a batch of videos is loaded for a training iteration, it comprises of (F+4)-frame ``videos'', which are split into 5 clips.
These (F+4) frames are randomly selected from the entire video.

\vspace{-0.1in}
\paragraph{Additional training details.} 
In each batch, we include 8 unique identities.
For each identity $\text{ID}_i$, the pull term (Equation~\ref{eq:pull}) comprises of 16 clips: 8 are self-reenactments, randomly sampled from the full set, and the remaining are cross-reenactments with $\text{ID}_i$ as the \emph{driving} identity.
These cross-reenactments can potentially show the same words being spoken by different target identities.
This is crucial: it allows the neural network to learn to to pull together videos based purely on the facial motion, regardless of the appearance of the video.
The push term (Equation~\ref{eq:push}) for $\text{ID}_i$ is composed of clips with the remaining 7 identities in the batch serving as driving identities (8 clips per driving identity).
Therefore, for each identity, 72 clips are included in a batch. 
The training is performed for 100,000 iterations, with Adam optimizer~\cite{kingma2014adam} and a learning rate of $1e^{-4}$.
During evaluation, one clip is sampled from each video and AUC is reported based on comparisons of each clip against the positive and the negative set.

\section{Evaluation}\label{sec:supp_evaluation}

\begin{table}[t!]
    \setlength\tabcolsep{5pt}
    \centering
    \scriptsize{
        \begin{tabular}{c|c}
        Clip duration F   &       AUC     \\\hline
        31         &      0.820          \\ %
        51         &      0.852     \\ 
        71         &      0.868        \\
        \end{tabular}
    }
    \vspace{0.1in}
    \caption{Ablation study for different values of F.}\label{table:clip-duration}
    \vspace{-0.2in}
\end{table}

\begin{table}[t!]
    \setlength\tabcolsep{5pt}
    \centering
    \scriptsize{
        \begin{tabular}{c|c}
            compression level  &       AUC     \\\hline
            Low (CRF 5) & 0.868 \\
            Medium (CRF 25) & 0.859 \\
            High (CRF 40) & 0.849
        \end{tabular}
    }\vspace{0.1in}
    \caption{Robustness to video distortions.}\label{table:compression-robustness}
    \vspace{-0.2in}
\end{table}

\begin{table}[t!]
    \centering
    \setlength\tabcolsep{5pt}
    \resizebox{0.8\columnwidth}{!}{
        \begin{tabular}{c|c|c|c}
        \scriptsize experiment name & \scriptsize positive set & \scriptsize negative set & \scriptsize AUC \\
        \hline
        \scriptsize \makecell{reference-driven cross\\reenactments in +ve set} &\scriptsize\makecell{driver = ref. ID;\\target = all IDs} & \scriptsize\makecell{driver = other IDs;\\target = ref. ID}& \scriptsize 0.83 \\
        \hline
        \multirow{2}{*}{\scriptsize scripted-set analysis} &\scriptsize\makecell{others utterances;\\driver = ref. ID} &\scriptsize\makecell{same utterance as test vid.;\\driver = other IDs} & \scriptsize 0.79\\
        \cline{2-4}
         &\scriptsize\makecell{others utterances;\\driver = ref. ID} &\scriptsize\makecell{other utterances;\\driver = other IDs} &\scriptsize 0.80
        \end{tabular}
    }
    \vspace{0.1in}
    \caption{Analysis with scripted subset.}\label{table:scripted-analaysis}
    \vspace{-0.2in}
\end{table}

We report additional ablation studies and robustness analyses in this section.
Specifially, we evaluate the impact of reducing the clip duration (Table~\ref{table:clip-duration}), the robustess of our trained model (trained on facevid2vid videos, at F = 71) to varying levels of compression (Table~\ref{table:compression-robustness}), and controlled analyses on scripted subset of our dataset (Table~\ref{table:scripted-analaysis}).

\paragraph{Impact of varying clip duration.} In Table~\ref{table:clip-duration}, we report the the results of our experiment with varying values of F, which is the number of frames provided to the network to make a prediction about the dynamic facial temporal identity signature.
The performance gained with increasing values of F diminishes at high values. 
We choose 71 frames as the default for most of our experiments.
For cases where shorter clips are desirable, such as for efficiency or for doing frequent verification, we observe that F=31 is plausible---with an AUC of 0.820.
Note that, as also discussed in Section~\ref{sec:dataset} we include a large set of scripted monologues in our dataset, which are crucial for a complete evaluation of avatar fingerprinting.
These video clips tend to be of a shorter duration (since subjects can recite only a few sentences at a time).
Therefore, we are constrained by the shortest-duration video clips in the dataset (98\% of the video clips are at least 71 frames long) and the largest value of F we experiment with is 71.

\begin{figure}
    \centering
    \input{figures/results/more_video_results}
\end{figure}

\paragraph{Robustness to distortions.}
We vary the quality of videos by compressing the test videos to three different levels of compression (color perturbations are observed with higher compression) and test the performance of our trained model (trained on face-vid2vid, f = 71). 
We observe a negligible performance drop---see Table~\ref{table:compression-robustness}.

\noindent \paragraph{Analysis with scripted subset (Table~\ref{table:scripted-analaysis}).}
In Table~\ref{table:scripted-analaysis}, we analyze the performance of our model over the scripted subset in the test set.
We want to assess the following:
\begin{enumerate}
    \item Are cross-reenactments driven by a given identity (``reference identity'') are closer to each other in the embedding space, as compared to those cross-reenactments where reference is the target for other drivers?
    \item In the scripted set, do embeddings for video driven by reference lie closer than the cross-reenactment videos where reference is driven by other driver speaking the same utterance and different utterances?
\end{enumerate}

The evaluation in such a controlled setting, which is only possible using our dataset, allows us to assess the clustering of the embedding space independent of the words being spoken.
We report the average AUC over the scripted test subset for when test-video utterance matches the cross-driven set (row 3 Table~\ref{table:scripted-analaysis}), and when it does not (row 4 Table~\ref{table:scripted-analaysis}). 
The result is slightly worse when the utterance of the cross-driven set (``negative class'') matches the test video, given the similar lip movements.

\paragraph{Additional results.} In Figure~\ref{fig:morevideos}, we show more samples similar to the ones shown in Figure~\ref{fig:dataset} from the main paper, for face-vid2vid generator (since this generator shows the best quality).
For each row of results in Figure~\ref{fig:morevideos}, we choose a reference identity, and a held-out set of reference self-reenacted videos for each of these identities.
Then, we report the average Euclidean distance of the following videos with respect to the held-out self-reenacted videos for the reference:
\begin{enumerate}[noitemsep,topsep=0pt,parsep=0pt,partopsep=0pt]
    \item a new self-reenacted video by the reference identity (not included in the held-out reference set) -- highlighted with a green border, 
    \item a cross-reenacted video where the reference identity is the driver -- highlighted with a green border, and
    \item two cross-reenacted videos where the reference identity is the target, driven by some other identity -- highlighted with a red and a blue border.
\end{enumerate}
Based on the reported distance values, we observe that videos where the reference identity is the driver are closer to the set of other self-reenacted videos driven by the reference identity and far from those where reference identity is the target to be driven by other identities.
This further confirms the ability of our model to fingerprint synthetic avatars based purely on facial motion, independent of the appearance of a synthetic talking-head video.

\end{document}